%% file: st_LID.tex
\documentclass[runningheads]{llncs}
\usepackage[T1]{fontenc}
\usepackage{booktabs}
\usepackage[misc]{ifsym}
\newcommand{\corr}{(\Letter)}
\usepackage{hyperref}       
\usepackage{url}            
\usepackage{amsfonts}       
\usepackage{microtype}      
\usepackage{xcolor}      
\usepackage{graphicx} 
\usepackage{multirow}
\usepackage{amsmath}
\usepackage{caption}
\usepackage{subcaption}
\usepackage{stackengine}

\input{math_commands}

\begin{document}

\title{Local Intrinsic Dimensionality of Ground Motion for Early Detection of Catastrophic Slope Failure}

\titlerunning{Local Intrinsic Dimensionality for Slope Failure Detection}

\author{Yuansan Liu\inst{1}\orcidID{0000-0003-3990-662X} \corr \and
James Bailey\inst{2}\orcidID{0000-0002-3769-3811} \and
Antoinette Tordesillas\inst{1}\orcidID{0000-0001-5497-292X}}

\authorrunning{Y. Liu et al.}

\institute{The University of Melbourne, Melbourne VIC 3053, Australia \email{\{yuansan.liu1,atordesi\}@unimelb.edu.au}
\and Monash University, Clayton VIC 3800, Australia \email{james.a.bailey@monash.edu}
}


\maketitle

\begin{abstract}
  Local Intrinsic Dimensionality (LID) has shown strong potential for anomaly detection in high-dimensional data, including landslide failure detection in granular media, where early and accurate identification of failure zones is crucial for effective geohazard mitigation. However, this task is still challenging due to the spatial correlations and temporal dynamics that are inherently present in surface displacement data. To address this gap, we propose a novel unsupervised framework called spatiotemporal LID ($st$-LID) that generalizes the LID for robust failure detection in landslide monitoring networks. Our approach introduces three key innovations: (1) Kinematic enhancement, incorporating velocity into the LID computation to capture instantaneous deformation rates and short-term temporal dynamics; (2) Bayesian spatial fusion, which aggregates LID values across spatial neighborhoods via Bayesian estimation, to embed spatial correlations and account for localized noise; and (3) Temporal modeling ($t$-LID), a new variant that characterizes long-term dynamics of displacement data, providing a robust temporal representation of displacement behavior. By unifying these components, $st$-LID identifies complex, multi-stage failure zones often overlooked by existing methods. Extensive experiments show that $st$-LID consistently outperforms state-of-the-art unsupervised baselines in detection precision and lead-time, providing a robust foundation for landslide early warning systems and targeted risk intervention to enhance community resilience and preparedness strategies.
\end{abstract}


\section{Introduction}
 Local Intrinsic Dimensionality (LID) has recently emerged as a robust geometric tool for characterizing anomalous behavior in high-dimensional datasets. By quantifying the expansion rate of a sample’s neighborhood, LID has demonstrated significant successes across various scenarios such as intrusion detection for IoT networks \cite{gorbett2022local}, identifying adversarial attacks on neural networks \cite{ma2018characterizing}, and outlier detection for high-dimensional data \cite{anderberg2024dimensionality}. Recent advances in granular media mechanics have further demonstrated that LID can detect impending failures in small scale laboratory tests by identifying shifts in the latent dimensionality of deformation \cite{zhou2021local,Tordesillas:2022wb}.
 
 At larger scales, such failures manifest as some of the most devastating natural hazards. Among these, catastrophic landslides are becoming increasingly frequent and severe due to climate change, seismic activity, and human intervention \cite{Sim:2022vi,abd496edaab54d908839a15be197653a,HAQUE2019673}. They can occur suddenly with little warning, often resulting in significant infrastructure damage, loss of life, and long-term economic disruption \cite{dai2002landslide,Lacroix:2020ui,Sim:2022vi}. Identifying anomalies in such systems is particularly challenging because they involve multi-regime transitions driven by distinct physical processes. It is rare for a single metric to track the evolution of a multi-event process across spatial and temporal scales. This underscores the importance of timely and accurate identification of high-risk areas to enable early intervention, save lives, and reduce property loss. 
 
 Existing works commonly apply traditional machine learning and statistical techniques, such as $K$-means clustering \cite{tordesillas2021spatiotemporal} and Empirical Dynamic Quantiles (EDQs) \cite{pena2019empirical,tordesillas2024augmented}, to monitoring data on surface displacement in the precursory failure regime, which has been proved to be one of the most direct indicators for identifying the location and timing of slope collapses \cite{intrieri2018maoxian,zhou2022pinpointing}. However, the high dimensionality and inherent spatiotemporal characteristics of such data often challenge the traditional detection methods which are built upon uni-variate time series analysis. As a result, the original informative, multidimensional data is frequently reduced to single dimensional data, leading to significant information loss and potentially inaccurate detection, in both spatial and temporal aspects. Therefore, these existing works (1) \textit{fail to fully exploit both spatial and temporal dependencies inherited in the spatiotemporal displacement data}, (2) \textit{fail to identify the sudden changes or accelerations in displacement over time}, (3) \textit{fail to capture the coherent pathways of regime transitions}. Subsequently, these limitations result in an inability to distinguish between steady-state motion and the subtle precursors of a dynamical regime shift toward catastrophic failure, leading to inaccurate detection, missing opportunities for timely intervention. 
 
 To address the challenge in learning with spatiotemporal data, we extend earlier studies of LID \cite{zhou2021local} to establish a LID-based framework for dynamic detection of high-risk areas most prone to slope failure at large field scales, by accounting for the outlyingness of kinematic behavior across multiple \textbf{spatial and time scales}. We first add the velocity ($v_t$) at each time stamp into target sample, allowing the algorithm to leverage the variability of the displacement and learn the short-term temporal dependence. Incorporating velocity is crucial, as many landslide early warning frameworks rely on velocity thresholds to trigger alerts and guide response actions \cite{CROSTA20021557,YU2025105232}. Secondly, we apply Bayesian estimation to aggregate LID results in a small neighborhood, accounting for spatial dynamics of the query points. Thirdly, we propose $t$-LID to measure the outlyingness of a target sample with respect to its $t$ nearest temporal neighbors (historical records), enabling it to capture the long-term dynamics from the entire time series data. Finally, we integrate two LID-based scores to highlight the points that are outlying with respect to both spatial and temporal neighborhoods. 
 \begin{figure}[ht]
     \centering
     \includegraphics[width=0.7\linewidth]{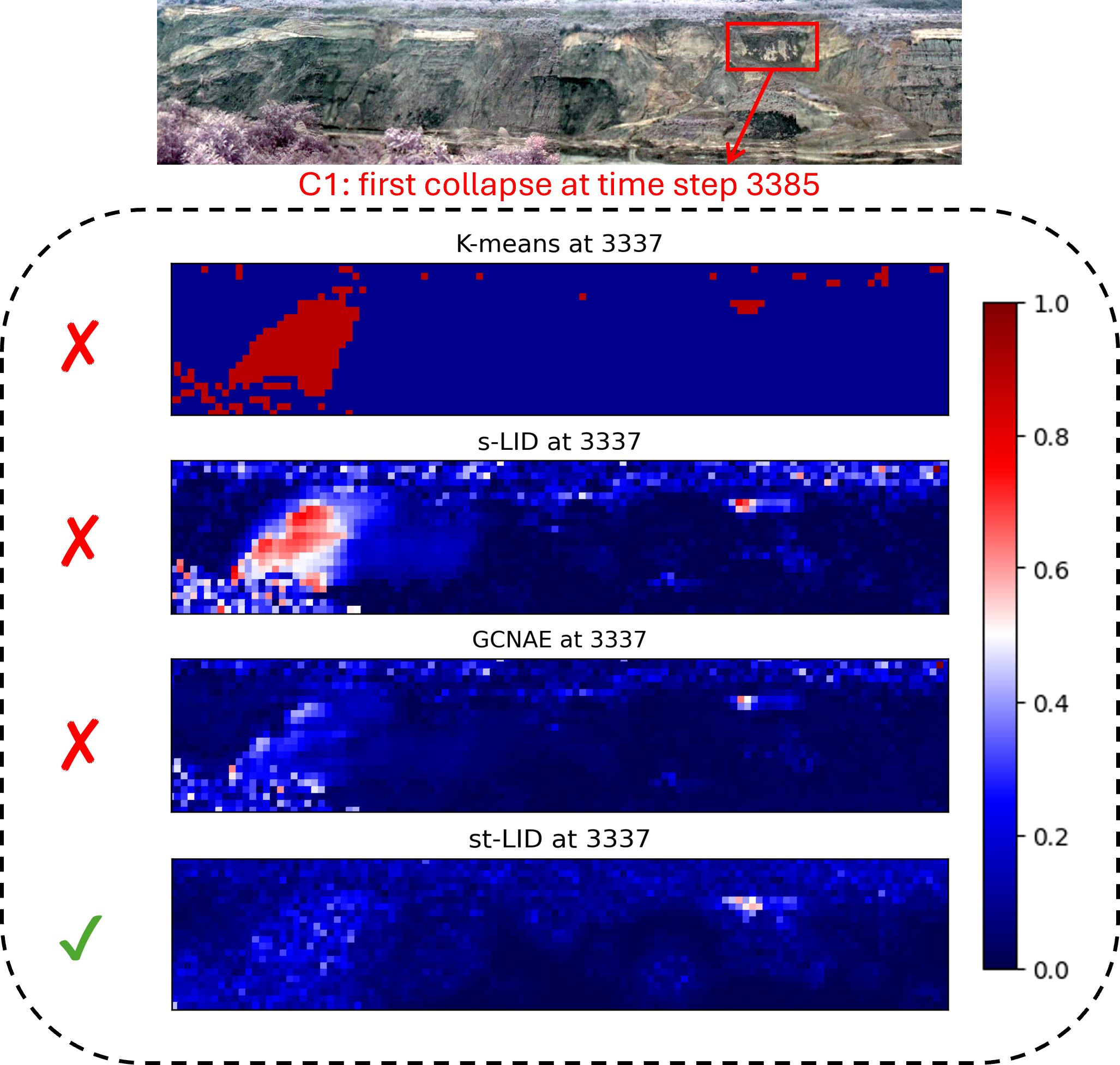}
     \caption{Detection of location and order of successive failures in distinct sites of a slope. Top: The first failure event $C1$ (red frame) occurred at $t=3385$. Bottom: Four methods detect impending failure locations at an earlier time step $t=3337$ (2 hours before collapse $C1$). Each monitoring point is colored based on the likelihood of failure at its location from 0 to 1. K-means produce binary results, 0 or 1. EDQs highlight points representative of the dynamics against the lower bound 0 for background. $s$-LID correctly identifies area of potential collapse but with extra noise. \textit{\textbf{Only the proposed st-LID can detect and isolate the location of the first failure event $C1$ that is relevant at this time step.}}}
     \label{example}
 \end{figure}
 Specifically, we apply a sigmoid function to normalize both scores into probabilistic values and integrate them via multiplication to emphasize samples where both scores are high. This integration not only reduces false positives but also enables earlier detection of hazardous shifts, providing actionable insights well before visible damage occurs. We name the final score \textbf{s}patio\textbf{t}emporal \textbf{LID} ($st$-LID). Unlike traditional metrics that struggle to describe transitions across different physical processes, st-LID serves as a unified geometric measure which is able to track the evolution of multi-dynamical-regime processes. By combining spatial outlyingness with temporal shifts into a calibrated $[0, 1]$ anomaly score, the method produces interpretable visions of coherent failure pathways. This allows for the early identification of regime transitions in a strictly unsupervised manner, without the need for labels or prior model fitting. Experimental results demonstrate that the proposed $st$-LID can identify the actual landslide location in both single and multiple-failure cases with accuracy and efficiency. This highlights the practical value of $st$-LID in real-world landslide early warning systems, with the potential to save lives and protect critical infrastructures.

 Our contributions can be summarized as:
 \begin{itemize}
     \item We propose a novel latent geometry-based method called $st$-LID for spatiotemporal outlier detection. This approach integrates three key components to capture both spatial correlations and temporal dynamics, enabling effective and efficient detection of landslide failures.

     \item We validate the proposed $st$-LID through extensive experiments on real-world datasets, demonstrating its competitive performance with respect to precision and lead-time of failure detection, and the computation efficiency of the algorithm.

     \item We propose a systematic failure detection procedure based on $st$-LID, showcasing its performance in early prediction of precise failure locations. This highlights the practicability of the $st$-LID in real-world landslide early warning systems, enabling timely interventions and enhanced risk management.
 \end{itemize}
 
\section{Background \& Related Works}
 \subsection{Landslide Monitoring and Statistical Solutions}
\begin{figure}[ht]
     \centering
     \includegraphics[width=0.75\linewidth]{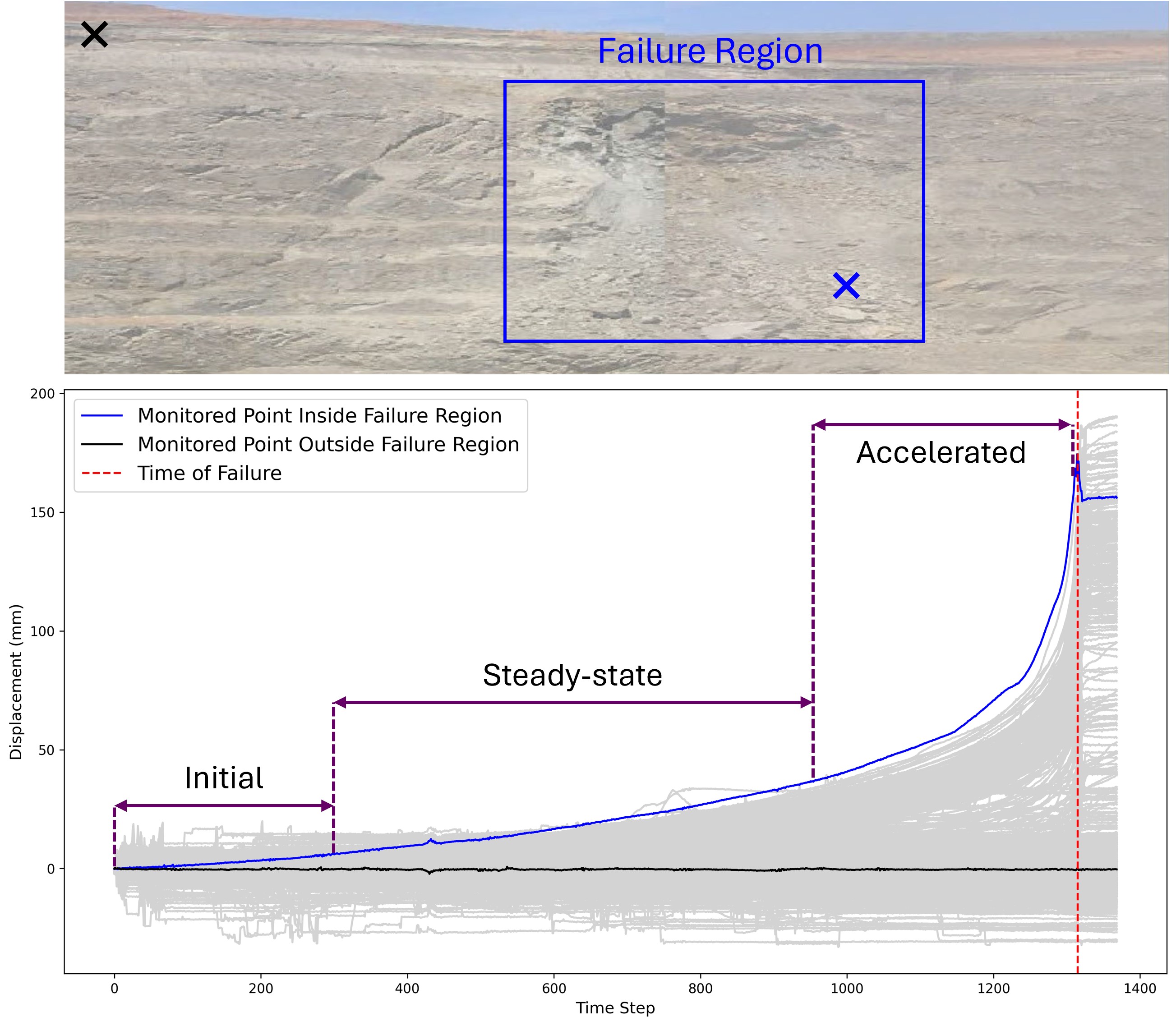}
     \caption{Example of a real-world slope Failure (Top) and the displacement time series of each monitored point (Bottom). The failure (a collapse advised by domain expert) is located inside the blue frame. Each time series at the bottom represents the displacements of a monitored point along with the time. Two monitored points (inside and outside the failure region) are marked via `$\times$' with different colors in the site map (their corresponding time series are highlighted with the same colors). The red dashed line indicates the time of failure (ToF).}
     \label{landslide}
  \end{figure}

  Surface displacement is a primary indicator of impending slope failure.\cite{intrieri2018maoxian,doi:10.1126/sciadv.adq9399}. Figure \ref{landslide} shows a typical example of the displacement time series corresponding to a catastrophic slope collapse. Notably, the displacement of points in the failure region manifest three stages of landslide creep (initial, steady-state and accelerated). 
  In the lead-up to collapse, ground motion exhibits a distinct spatiotemporal clustering dynamics, with high-risk zones increasingly differentiating themselves from more stable areas of the slope. These patterns highlight the potential for using machine learning and data mining techniques to enable the automatic early detection of slope failure based on surface displacement data. Traditional detection relies on statistical clustering (e.g., K-means, LOF), which identifies high-risk regions but often yields noisy, binary results that ignore temporal dynamics\cite{tordesillas2021spatiotemporal,zhou2022pinpointing}. Alternatively, Empirical Dynamic Quantiles (EDQs) detects failure locations by analyzing high-percentile displacement behaviors \cite{tordesillas2024augmented}. However, EDQs process sensors independently, lacking the spatial regularization necessary to handle measurement noise, leading to sparse and disjointed detections.

 \subsection{Local Intrinsic Dimensionality}
  Rather than analyzing raw values, Local Intrinsic Dimensionality (LID) exploits latent data properties to identify outliers \cite{houle2017local,zhou2021local,anderberg2024dimensionality}. LID measures the expansion rate of a sample’s neighborhood, where a high value indicates the sample is outlying relative to its neighbors. For a sample $x$ and its cumulative distance distribution $\mathcal{F}(r)$ to its neighbors, LID is defined as:
  \begin{equation}
      \text{LID}(x) = \lim_{r\rightarrow 0} \text{LID}_{\mathcal{F}(r)} = r\frac{\mathcal{F}'(r)}{\mathcal{F}(r)}.
  \end{equation}
  
  In practice, the Maximum Likelihood Estimator (MLE) is used to approximate LID over $s$ nearest neighbors. For landslide detection, the baseline $s$-LID computes this at fixed time steps based on displacement values \cite{zhou2021local}:
  \begin{equation} \label{slid}
      s\text{-LID}(\vx_t) = -\left( \frac{1}{s}\sum_{i=1}^s \log \frac{dist_i(\vx_t)}{dist_s(\vx_t)} \right)^{-1},
  \end{equation}
  where $dist_i(\vx_t)$ is the Euclidean distance of $\vx_t$ to its $i^{th}$ nearest neighbor with displacement value of $x_t^i$: $dist_i(\vx_t) = |x_t^p - x_t^i|$. While $s$-LID provides more granularity than K-means, it fails to leverage temporal dynamics and remains susceptible to sparsely distributed high-risk noise.

 \subsection{Deep Learning Models on Spatiotemporal Analysis}
  The advancements in deep learning have introduced powerful tools for learning spatial information in sensor networks. State-of-the-art models like Graph Convolution Networks (GCNs) or Graph Attention Networks (GATs) are used to extract the spatial topology between monitored points, identifying anomalies through reconstruction error. Recent works also combine these spatial models with the temporal modules such as LSTM or Transformer to jointly learn the spatial and temporal dynamics for environmental forecasting tasks \cite{velickovic2018graph,ding2019deepgcn,wang2024lstm,sun2025flexireg}. 
  
  While these deep learning approaches are able to achieve state-of-the-art performance, they typically require extensive labeled datasets for \textbf{supervised or semi-supervised} training which are extremely sparse in real-world. On the other hand, our proposed $st$-LID offers a \textbf{strictly unsupervised} solution, enabling online detection with a stream of incoming data without any training or tuning. Additionally, in contrast to these ``black-box'' models, $st$-LID maintains the mathematical interpretability of the LID expansion model while explicitly incorporating physical kinematics and Bayesian uncertainty to achieve state-of-the-art performance without the need for prior failure labels.

\section{Methodology}
 To address the limitation of existing methods on learning with spatiotemporal data and improve its practicability in landslide application, we propose a novel spatiotemporal measurement to leverage both spatial correlation and temporal dynamics inherent in the displacement data.

 \subsection{Leveraging Velocity in LID Calculation}
  Firstly, to introduce short-term temporal dependency, we extend the data sample into a vector consists both displacement value and velocity $v_t^p = x_t^p - x_{t-1}^p$ at monitored point $p$ and time step $t$, $\vx_t = \langle x_t^p, v_t^p\rangle$. And the distance of a queried sample $\vx_t$ to its $i^{th}$ nearest neighbor at monitored point $i$ becomes:
  \begin{equation} \label{dist}
     dist_i(\vx_t) = \sqrt{(x_t^p - x_t^i)^2 + (v_t^p - v_t^i)^2}.
  \end{equation}

  Applying this new distance calculation to equation (\ref{slid}), we get the $s\text{-LID}(\vx_t)$ that incorporates both displacement value and velocity at given time step. This inclusion allows the metric to perceive the instantaneous shift in dynamical behavior, identifying the transition between different motion regimes. Consequently, it effectively mitigates the bias introduced by samples exhibiting high displacement values but remain relatively stable.

\begin{figure}[h]
     \centering
     \includegraphics[width=0.85\linewidth]{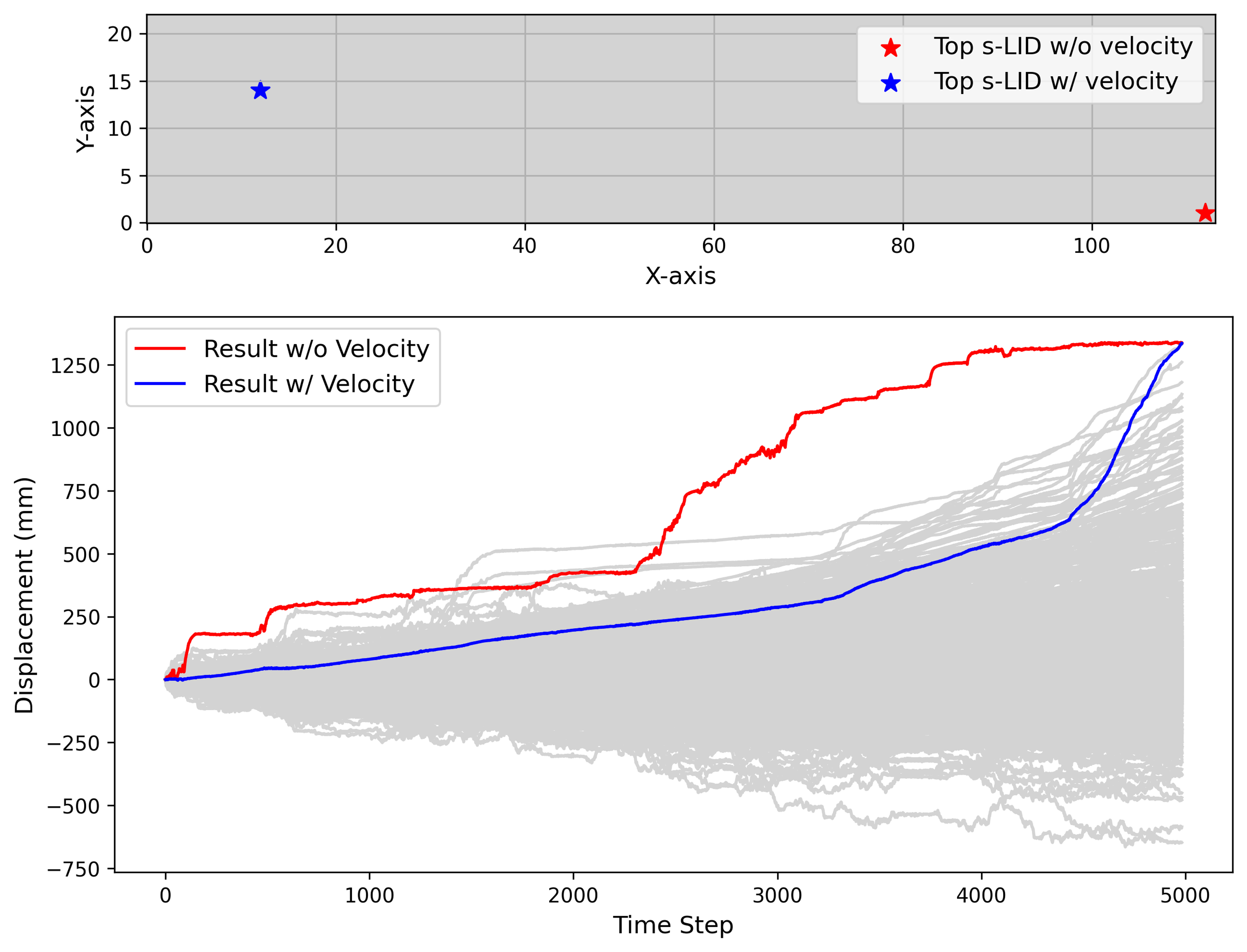}
     \caption{Example high-risk areas and corresponding time series with \& without velocity in $s$-LID computation. Top: Site map at last time step ($t=4983$), with two monitored points highlighted by top $s$-LID values, representing the high-risk locations. {\color{red}$\star$} is the $s$-LID without velocity, {\color{blue}$\star$} is the $s$-LID with velocity. Bottom: Displacement time series for each monitored point, with the same color.}
     \label{slid_v}
  \end{figure}

  Figure \ref{slid_v} depicts the effect of adding velocity with an real-world example: In the site map at top, different monitored points are selected as high-risk candidates ({\color{red}$\star$} and {\color{blue}$\star$}) according to the largest $s$-LID values at the last time point ($t=4983$) when queried samples with or without the velocity. The corresponding time series at bottom illustrates the different temporal dynamics between these two monitored points, reflecting the aforementioned bias when using only displacement values for $s$-LID computation. Note that the actual failure region is around the monitored point {\color{blue}$\star$}. This fact justifies our enhancement of incorporating the velocity component. 

 \subsection{Spatial Fusion of LID}
  Although $s$-LID measures the query point's characteristic against its spatial neighbors, it finds the nearest neighbors in the kinematic space, without involving the spatial dependencies in physical space. To leverage the spatial relationships of all monitored points during the LID calculation, inspired by Bayesian LID algorithm \cite{joukhadar2024bayesian}, we integrate the $s$-LID results in a small neighborhood of the query point, enhancing the LID score with actual spatial dynamics. 
  \begin{figure}[ht]
      \centering
      \includegraphics[width=0.5\linewidth]{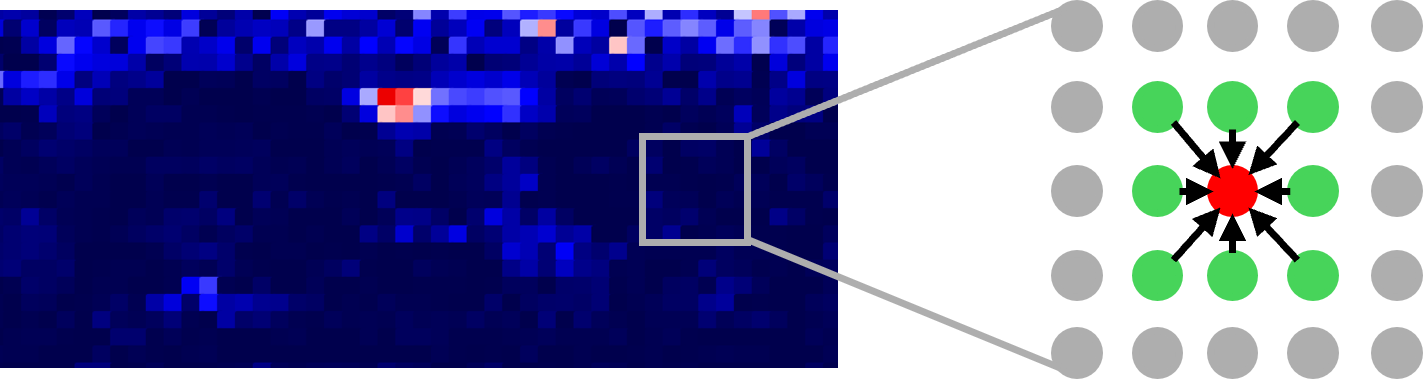}
      \caption{Spatial fusion of $s$-LID. Red dot in the middle is the query point and its neighbors are colored with green. The final LID value of the query point is calculated based on its own observation and the pre-calculated $s$-LID of these neighbors at last time step.}
      \label{spatiallid}
  \end{figure}
  Figure \ref{spatiallid} depicts the underlying idea, where the query point (red) is updated based on its neighbors (green).

  Given a region with pre-calculated $s$-LID values at time step $t-1$. For each query point $\vx_t$, let $\{f^i_{t-1}(l) \}_{i=1}^k$ be the $k$ neighbors' prior densities that are built from last time step $s$-LID values. With a set of weights $\vw = \{w_1, \cdots, w_k \}$ summing up to $1$, we can combine these priors with logarithmic pooling:
  \begin{equation} \label{log_pool}
      f^{pool}_t(l|\vx_t) := \left(\int_0^{\infty} \prod_{i=1}^k (f^i_{t-1}(l))^{w_i}dl \right) \prod_{i=1}^k (f^i_{t-1}(l))^{w_i}.
  \end{equation}

  To simplify the calculation, we can assume Gamma densities $f^i = Ga(\alpha_i, \beta_i), i=1,\cdots,k$, and equation (\ref{log_pool}) becomes a Gamma density as well: 
  \begin{equation} \label{ga_pool}
      f^{pool} = Ga(\alpha_p = \sum_{i=1}^k w_i\alpha_i, \beta_p = \sum_{i=1}^k w_i\beta_i).
  \end{equation}
  Here, we apply Gaussian kernel to create weights that account for the physical proximity of the query point and its neighbors: 
  $$w_i = \frac{\exp\left(-\frac{d_e(p, i)^2}{2\sigma^2}\right)}{\sum_{j \in k} \exp\left(-\frac{d_e(p, j)^2}{2\sigma^2}\right)}.$$
  Where $d_e(p, i(j))$ is the Euclidean distance between query point $p$ and its $i(j)^{th}$ neighbor. This weight ensures closer neighbors have higher influences of the pooling results. Then, the parameters of the Gamma density can be derived from the weighted mean and variance of the neighbors' $s$-LID values:
  \begin{equation}
      \mu = \sum_{i=1}^k w_i s\text{-LID}(\vx_{t-1}^i), \quad
      \sigma^2 = \sum_{i=1}^k w_i \left(s\text{-LID}(\vx_{t-1}^i)-\mu\right)^2, \quad
      \alpha_p = \frac{\mu^2}{\sigma^2}, \quad \beta_p = \frac{\mu}{\sigma^2}. \label{prior}
  \end{equation}
  Where $s$-LID($\vx_{t-1}^i$) is the $s$-LID value of the $i^{th}$ neighbor of query point at $t-1$.
  
  Once we get the current observation of query point, we can estimate its likelihood contribution via the density ratio within the $k$-neighborhood:
  \begin{equation} \label{obs}
      \alpha_o = k, \quad \beta_o = -\sum_{i=1}^k \log \frac{dist_i(\vx_t)}{dist_k(\vx_t)}.
  \end{equation}
  
  Finally, we can get the posterior estimation of the query point's $s$-LID with parameters from equations (\ref{prior}) and (\ref{obs}) following Gamma conjugacy:
  $$s\text{-LID}^* = \frac{\alpha_p+\alpha_o}{\beta_p+\beta_o}.$$
  This approach can effectively leverage spatial correlation in geophysical monitoring data to produce more robust LID estimates, ensuring geometric coherency and spatial continuity in LID calculation.

 \subsection{Temporally Informed LID}
  Thirdly, to learn the long-term dynamics of the time series data, we propose a temporal version of LID namely $t$-LID. Same as $s$-LID, $t$-LID is also estimated based on the distance expansion rate. Rather than spatial neighborhood, $t$-LID focuses on temporal neighborhood of the target sample, on another word, the historical values of the given sample until the analyzing time step $t$. Notice that, the raw displacement values are highly non-stationary across the temporal dimension. To mitigate the outlyingness contributed by such non-stationarity, we compute $t$-LID solely on the velocity values. This method also aligns with the common practice for de-trending non-stationary time series \cite{mills201971}. Hence, given a time series of samples at a fixed monitored point $l$: $\vx_{1:t} = \{\vx_1, \vx_2, \cdots, \vx_t\}, \vx_t = \langle v_t^l\rangle$, we can write $t$-LID of the queried sample at current time step $\vx_t$ as:
  \begin{equation}
     t\text{-LID} (\vx_t) = -\left( \frac{1}{t}\sum_{i=1}^t \log \frac{dist_i(\vx_t)}{dist_t(\vx_t)} \right)^{-1},
  \end{equation}
  where $t\geq 2$ and $dist_i(\vx_t) = |v_t^l - v_i^l|$ is the distance of the queried sample $\vx_t$ at current time step $t$ to its $i^{th}$ nearest neighbor at a historical time step $i$. 

  Similar to $s$-LID, which evaluates the outlyingness of a queried sample based on its spatial neighbors, $t$-LID measures the outlyingness of the queried samples against its own past records, enabling effective outlier detection within time series data. This temporal perspective allows $t$-LID to capture the dynamic changes along with the time, identifying the significant temporal deviations that may relate to the instability and emerging risks in a landslide process.

 \subsection{Spatiotemporal LID}
  Until now, we get two LID measurements: $s$-LID that quantifies the spatial outlyingness, and $t$-LID that captures the temporal deviations. At the final step, we propose a novel approach that integrates them into a unified spatiotemporal outlier detection framework. This integration aims to leverage the strengths of both methods on spatial and temporal outlier detection, addressing the limitations of traditional methods that consider these aspects in isolation. 

  To achieve an appropriate integration, we begin with the scale unification. Since spatial and temporal LID operate on different feature spaces and exhibit different score distributions, they are not directly comparable. We apply a sigmoid function $\sigma(x) = \frac{1}{1+e^{-x}}$ to map raw LID values into a normalized range $[0, 1]$. This step acts as a baseline calibration, mapping the local intrinsic dimensionality into a consistent outlier likelihood across varied monitoring sites. To ensure that only points exhibiting significant outlyingness in both spatial and temporal perspectives are highlighted, a multiplicative fusion is applied:
  \begin{equation}
     st\text{-LID}(\vx_t) = \sigma(s\text{-LID}(\vx_t)) \times \sigma(t\text{-LID}(\vx_t)).
  \end{equation}
  This multiplicative approach functions as a soft ``AND'' gate, effectively suppressing independent noise, such as localized sensor errors or transient temporal fluctuations, by amplifying signals where spatial clustering and temporal acceleration converge. Additionally, it also functions as a unified geometric measure designed to track multi-regime transitions. With $s$-LID captures the spatial outlyingness and $t$-LID identifies temporal shifts, the multiplication amplifies the signal only when these two distinct physical signatures align. This alignment marks the critical transition from a steady-state creep regime to an accelerated failure regime. While more complex fusion methods (e.g., rank-aggregation) could model specific dependencies, the multiplicative strategy provides a computationally efficient and interpretable precursor signal for real-time early warning.

 \subsection{Real-World Failure Detection via $st$-LID}
  The proposed $st$-LID is designed to identify the outliers in displacements data with precision and efficiency. However, it remains an open question to apply it in real-world scenario. In this section, we propose a systematic failure detection procedure to enable the practical deployment of $st$-LID.
  
  Assuming a continuous monitoring system based on $st$-LID scores at each monitoring time step, we detect a potential failure region around the monitored point $\vx^* \in \mathbb{R}^2$ at time step $t$ if for $n$ consecutive time steps $\{t-n+1, t-n+2 \cdots, t\}$ the following conditions stand:
  \begin{equation}\label{detect}
      arg \max_{\vx_t \in MP} st\text{-LID}(\vx_t) = \mathcal{B}_{\epsilon}(\vx_t^*); \quad st\text{-LID}(\vx_t^*) \geq 0.5\text{ .}
  \end{equation} 
  Where $MP$ is the set of all monitored points, and $\mathcal{B}_{\epsilon}(\vx_t^*) = \{\vy \in \mathbb{R}^2 \mid \| \vy-\vx_t^* \|<\epsilon \} \text{, } \epsilon>0\in\mathbb{R}$ is a small open ball around $\vx_t^*$ to allow noise related fluctuation.

  This procedure follows the thresholding of the failure ($\geq 0.5$) and ensures the convergence of the detection (small open ball). The choice of $n$ can be decided by the domain experts, acting as a trade off between sensitivity and fidelity. 
 
\section{Experiments}

 \subsection{Datasets}
  The datasets consist surface displacement measurements data from three real-world operating mine sites which experienced fatal collapses (failures). Due to confidential policy, we name these datasets as $M_1$, $M_2$, and $M_3$ respectively, without revealing their actual locations and names.
  \begin{itemize}
      \item $M_1$: Contains 2622 monitored points with displacement recorded in 8000 time steps, and there are two ground truth collapses happened at two different locations at $t=3385,7320$ respectively, we refer to them as $C1$ and $C2$.

      \item $M_2$: Contains 5844 monitored points with displacement recorded in 6000 time steps, and the single ground truth collapse happened at $t=5264$.

      \item $M_3$: Contains 6624 monitored points with displacement recorded in 6500 time steps, and the single ground truth collapse happened at $t=5806$.
  \end{itemize}
  The actual time interval between two consecutive time steps is $2.5$ minutes in $M_1$, and $6$ minutes in $M_2$ and $M_3$. More details of these datasets including the visualization of the displacement time series are provided in the Appendix.

 \subsection{Experimental Setup} 
  The target of this work is to pinpoint the real failure locations at early time step, therefore, we evaluate the methods using two metrics:
  \begin{itemize}
      \item \textit{Precision} ($Prec.$) measures the proportion of correct detection results (failure/collapse actually happened at the detected locations) relative to all detections from the method: $Prec. = \frac{\text{Correct Detections}}{\text{All Failure Detections}}.$ Higher precision reflects less false alarms, indicating trustworthy detection.

      \item \textit{Lead-time} ($\Delta t$) measures the difference between the model detection time and the actual failure time, quantifying efficiency of the method: $\Delta t = t_{failure} - t_{detection}.$ Longer lead-time represents earlier detection of failure, enabling timely invention and reactions.
  \end{itemize}
  
  We compare the proposed $st$-LID with existing failure detection methods in industry including K-means and $s$-LID. We also include a cluster method: Density-based spatial clustering of applications with noise (DBSCAN), and a deep learning based SOTA anomaly detection model Graph Convolution Network Autoencoder (GCNAE). Due to the differences in output types across these methods, we unify the detection threshold for fair comparison. Specifically, in K-means and DBSCAN, all monitored points from high-risk class are regarded as failure detections. For GCNAE, we follow their original paper to get the anomaly score from the reconstruction error. The hyperparameters in all models are fixed across different dataset, detailed setup for these models is provided in the Appendix. The raw outputs of both GCNAE and $s$-LID are linearly converted to the range $[0,1]$, and the failure detections are those with a final score $\geq 0.5$. The precision score $Prec.$ are computed using the detection results at the actual time of failures (collapses).
  
  Regarding the lead-time $\Delta t$ computation, to avoid the impact of noisy results in baselines, we use the top $5$ results from each method, because it is the minimum detection numbers across all methods on all datasets. To be specific, these are the monitored points with: $5$ shortest distance to outlying cluster's centroid in K-means and DBSCAN and top $5$ scores for other methods. We record the first time when the selected points consistently fall within the actual failure regions (until the time of failure), and use it to compute $\Delta t$. We assign $\Delta t = 0$ for those methods cannot pinpoint the actual failure ahead of time.

  Note that we do not include the commonly used metrics \textit{Recall} because it requires a reliable count of all actual failures which is typically unavailable for real-world landslide scenarios. Moreover, the objective of this work is to develop manageable and high-confident early warnings for landslide failures, rather than to maximize the number of detected events at any false alarm burden. Thus, we focus on the \textit{Precision} and \textit{Lead-time} only.

 \subsection{Experimental Results}
  We provide both the qualitative and quantitative results in this section via the visual analysis and the aforementioned numeric metrics. The former provides intuitive insights into failure patterns, while the latter offers rigorous evaluation of performance.
 
   \begin{figure*}[ht]
      \centering
      \includegraphics[width=0.99\linewidth]{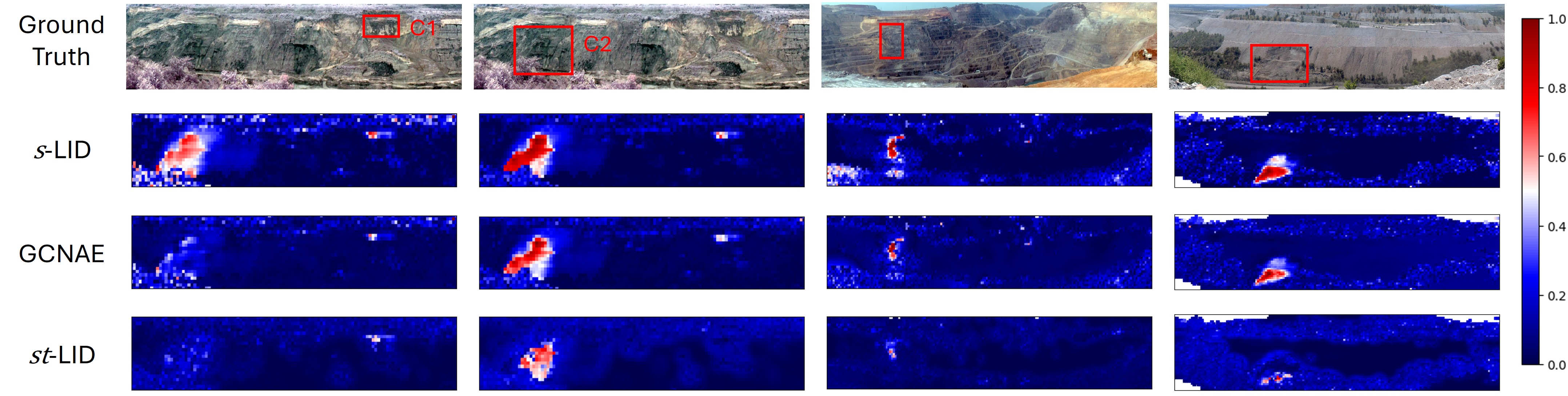}
      \caption{Visualization of the ground truth in 4 datasets (columns) and corresponding detection results from top $3$ methods (rows). Detection results are colored based on the likelihood of the failure (see color-bar for detailed range and order)}
      \label{vis}
  \end{figure*}
  
  Figure \ref{vis} shows the ground truth collapse locations (red rectangle frame, advised by the domain experts) along with the detection results from top $3$ methods (according to $Prec.$, visualizations of other methods are provided in Appendix). The proposed $st$-LID can precisely pinpoint the actual collapse locations in all four cases while other methods picking up both collapsed locations and the noisy results in most cases.

  \begin{table*}[ht]
    \centering
    \caption{Precision Score $Prec.$ ($\uparrow$) \& Lead-time $\Delta t$ ($\uparrow$) results from all methods. $\uparrow$ means higher values are better. Best performance are in bold format. Lead-time are shown in both raw time steps and their equivalent durations in days ($d$).}
    \begin{tabular}{c|c|c|c|c|c}
    \hline
    Metrics & Methods & $M_1$ ($C1$) & $M_1$ ($C2$) & $M_2$ & $M_3$ \\ \hline\hline
    \multirow{5}{*}{$Prec.$} & K-means & 0.044 & 0.940 & 0.128 & 0.999 \\ \cline{2-6}
     & DBSCAN & 0.047 & 0.557 & 0.298 & 0.705 \\ \cline{2-6}
     & $s$-LID & 0.045 & 0.945 & 0.725 & 0.896 \\ \cline{2-6}
     & GCNAE & 0.333 & 0.947 & 0.857 & 0.975 \\ \cline{2-6}
     & $st$-LID (ours) & \textbf{1.000} & \textbf{1.000} & \textbf{1.000} & \textbf{1.000} \\ \hline\hline
    \multirow{5}{*}{$\Delta t$} & K-means & 0 & 220 (0.4 $d$) & 0 & 427 (1.8 $d$) \\ \cline{2-6}
     & DBSCAN & 0 & 1351 (2.4 $d$) & 214 (0.9 $d$) & 1096 (4.6 $d$) \\ \cline{2-6}
     & $s$-LID & 0 & 1489 (2.6 $d$) & 1383 (5.8 $d$) & 1142 (4.8 $d$) \\ \cline{2-6}
     & GCNAE & 0 & 1516 (2.6 $d$) & 1473 (6.1 $d$) & 1184 (4.9 $d$) \\ \cline{2-6}
     & $st$-LID (ours) & \textbf{80} (0.1 $d$) & \textbf{1726} (3.0 $d$) & \textbf{1760} (7.3 $d$) & \textbf{1214} (5.1 $d$) \\ \hline
    \end{tabular}
    \label{res}
  \end{table*}
  
  The detailed precision scores and lead-time are listed in Table \ref{res}, from which we can observe that the proposed $st$-LID consistently achieves perfect precision score, indicating its ability to pinpoint the failure region in a large area, which is valuable for landslide warning system. Furthermore, $st$-LID can also produce timely predictions well before the actual time of failure in all these cases, demonstrating its effectiveness and efficiency in landslide early warning tasks.

 \subsection{Sensitivity Analysis}
  A sensitivity analysis is conducted using $M_2$ to evaluate the impact from the nearest-neighbor parameter $s$. We report the precision score under different parameters to compare how the performance changes with corresponding parameter. Table \ref{sen_ana} shows the precision score with different parameters.
  \begin{table}
      \centering
      \caption{Sensitivity Analysis on parameter $s$.}
      \begin{tabular}{c|c|c|c|c|c}
         \hline
         Parameter  & $s=500$ & $s=1000$ & $s=2000$ & $s=5000$ & $s=N-1$\\ \hline\hline
          $Prec.$ & 0.500 & 0.583 & 0.667 & 0.800 & 1.000\\\hline
      \end{tabular}
      \label{sen_ana}
  \end{table}
  It is clear that performance improves as $s$ increases, with $s=N-1$ giving the best precision for all three cases. This suggests that, for these field-scale displacement datasets, using a broader neighborhood is more robust than using a small local neighborhood. Smaller values can be too local and may respond more strongly to noise, local artifacts, or small deformation fluctuations rather than the coherent failure-region signal.

 \subsection{Ablation Study} \label{ablation}
  To quantify the contribution of each proposed enhancement,we conduct an ablation study on dataset $M_3$, systematically removing individual components and evaluating the impact on $Prec.$ and $\Delta t$ (Table \ref{abla}). The results indicate that velocity and spatial fusion primarily contribute on the $Prec.$, which aligns with their design to mitigate the impact of certain noise. The $t$-LID serves as the most critical part in this framework, enhancing both metrics significantly. By capturing long-term temporal dependencies, $t$-LID enables the framework to differentiate subtle pre-failure signals from steady-state motion, directly leading to the superior lead times observed in our primary results.
 
  \begin{table*}[ht]
    \centering
    \caption{Ablation Study on $M_1$ ($C1$) and $M_3$.}
    \begin{tabular}{cc|c|c|c|c}
        \hline
        \multicolumn{2}{c|}{Components} & $s$-LID w./o. $v$ & $s$-LID w. $v$ & Spatial Fusion & $st$-LID \\ \hline\hline
        \multicolumn{1}{c|}{\multirow{2}{*}{$M_1$ ($C1$)}} & $Prec.$ & 0.045 & 0.149 & 0.273 & \textbf{1.000} \\ \cline{2-6} 
        \multicolumn{1}{c|}{} & $\Delta t$ & 0 & 0 & 0 & \textbf{80} \\ \hline\hline
        \multicolumn{1}{c|}{\multirow{2}{*}{$M_3$}} & $Prec.$ & 0.795 & 0.896 & 0.927 & \textbf{1.000} \\ \cline{2-6} 
        \multicolumn{1}{c|}{} & $\Delta t$ & 1141 & 1142 & 1142 & \textbf{1214} \\ \hline
    \end{tabular}
    \label{abla}
  \end{table*}

 \subsection{Multi-failure Analysis}
\begin{figure}[ht]
      \centering
      \includegraphics[width=0.75\linewidth]{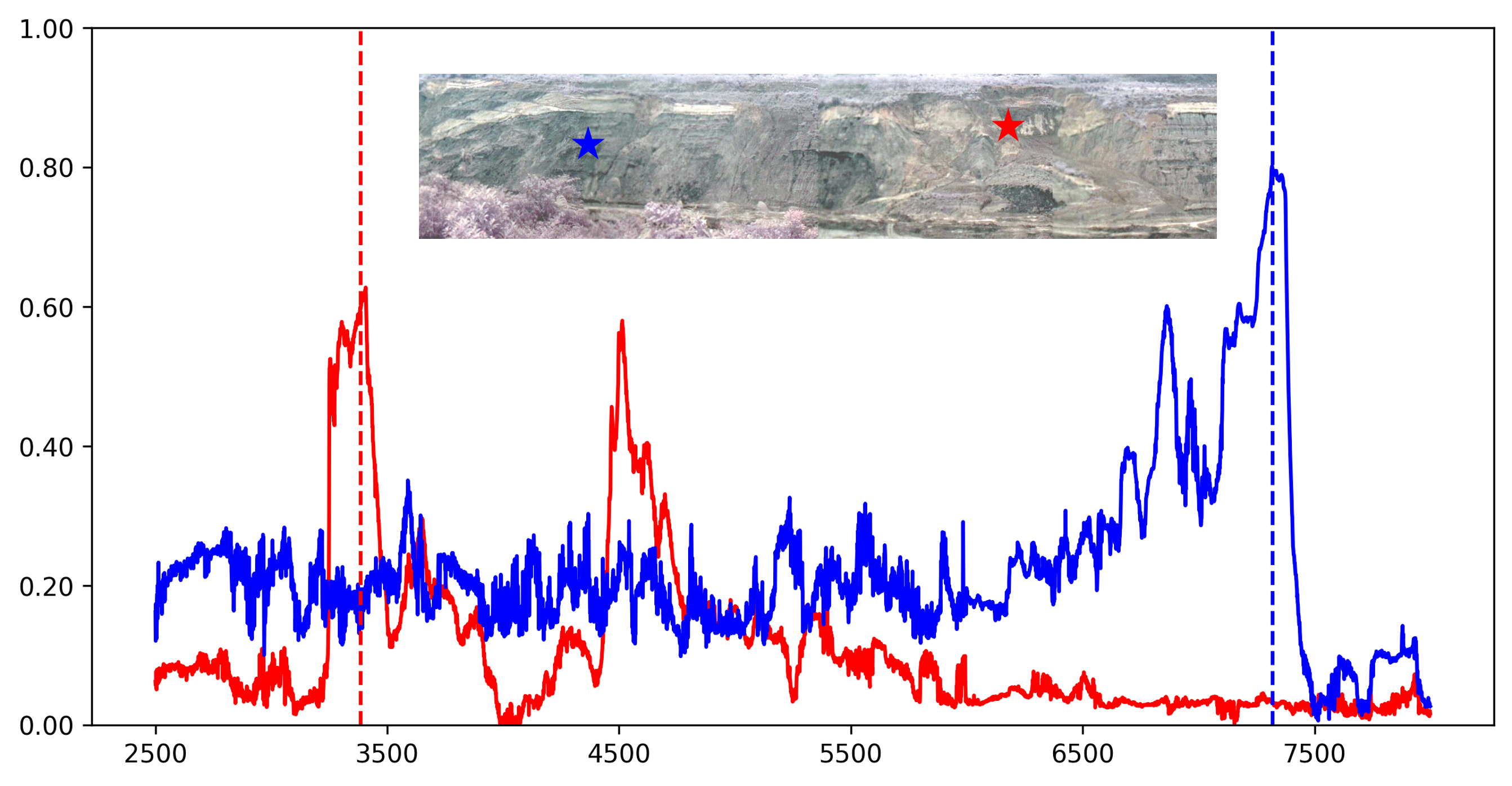}
      \caption{Evolution of $st$-LID scores across multiple dynamical regimes in $M_1$. The red and blue curves represent the scores for distinct regions on the slope (red and blue stars). The ToF of $C1$ and $C2$ are highlighted with dashed lines.}
      \label{multi}
  \end{figure}

  To further analyze the behavior of $st$-LID across multi-event processes, we demonstrate the evolution of the score for the two distinct failures in $M_1$. Figure \ref{multi} shows the detailed results of this analysis, with red and blue series represent one location in the $C1$ and $C2$ respectively (red and blue stars). From the figure, we can see $st$-LID successfully identifies the transition into the accelerated failure regime for both events, as evidenced by the clear peaks at their respective failure times (dashed lines). This underscores its ability to track dynamical regime shifts without the interference of background noise or stationary displacement signals.

  Notably, the red curve exhibits a secondary surge around $t = 4500$. While this specific interval was not initially identified as a standalone failure in the ground truth labels, subsequent geotechnical analysis of the corresponding deformation and coherence confirms it as a secondary sequence of collapses. This highlights a significant ability of $st$-LID to track these sub-critical regime transitions and successive physical failures with high fidelity.

 \subsection{Computation Time}
  Efficiency is one of the important factors in evaluation of landslide warning systems. A method loses its practical values if it requires excessive computation time, especially in real-time monitoring scenarios where timely alert are crucial for risk management. In this section, we compare the computation time of all methods.
  \begin{table}[h]
    \centering
    \caption{Computation time of each failure detection methods.}
    \begin{tabular}{c|c}
    \hline
    Methods & Computation Time \\ \hline\hline
    K-means & 0.192 seconds \\ \hline
    DBSCAN & 0.162 seconds \\ \hline
    $s$-LID & 0.649 seconds \\ \hline
    GCNAE & 3 hours (training) + 0.645 seconds \\ \hline
    $st$-LID (sequential) & 5.342 seconds \\ \hline
    $st$-LID (parallel) & 0.721 seconds \\ \hline
    \end{tabular}
    \label{time}
  \end{table}
  
  The results are listed in Table \ref{time}. Here, we record the computation time of each method detecting the potential failures at a single step. We provide two versions of $st$-LID because the $s$-LID integration and $t$-LID computation happens at each monitored point (each time series), thus, one can choose to sequentially compute it or parallel the computation. From the table, we can see except GCNAE, which takes significant time to train, all the other methods are efficient and only take seconds to finish. This fact further justify the practicability of $st$-LID in real-time monitoring and analysis of the landslide-prone area.

 \subsection{Real-World Failure Detection Simulation}
  Until now, we have demonstrated the effectiveness and efficiency of the proposed $st$-LID conceptually via multiple experiments. In this section, we simulate a real-time monitoring using $M_1$ dataset and demonstrate a practical failure detection procedure on $C1$ collapse.
  \begin{figure}[ht]
      \centering
      \includegraphics[width=0.8\linewidth]{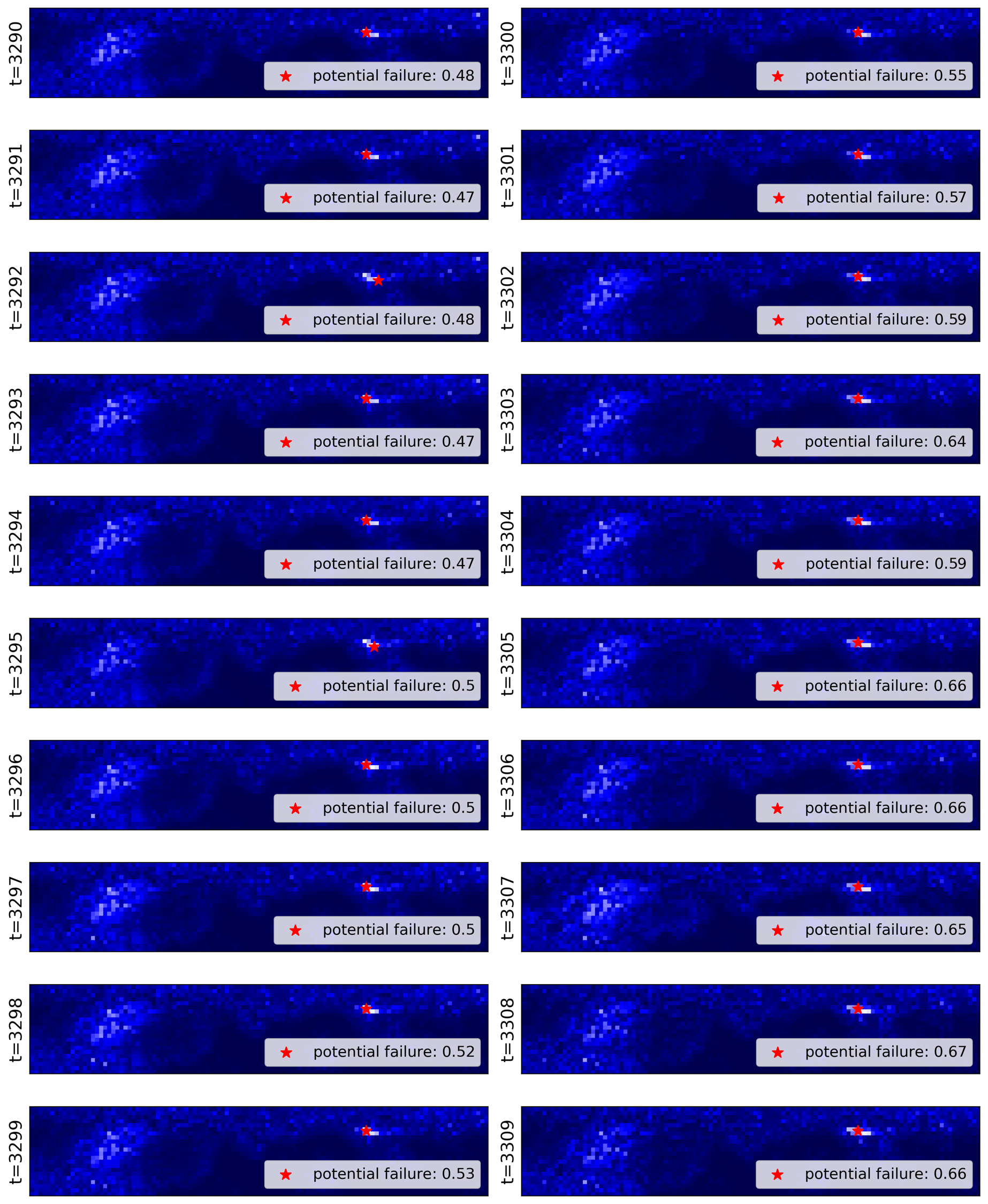}
      \caption{Simulation of real-world failure detection using $st$-LID. The monitored point with the highest $st$-LID is marked by {\color{red}$\star$}.}
      \label{det_sim}
  \end{figure}

  Applying the proposed detection procedure on $M_1$ ($C1$), we demonstrate a simulation of the failure detection, the details are shown in Figure \ref{det_sim}. Here, we use $n=10$ for higher fidelity considering the real-world impact of a failure detection. From the figure, we can see the target $\vx_t^*$ is fluctuating between two monitored points within a small region, following the small ball condition in equation (\ref{detect}). The target point $\vx_t^*$ does not meet the detection threshold at initial steps ($t\in[3290, 3294]$), but starts to consistently exceeds such threshold $st\text{-LID}(\vx_t^*) \geq 0.5$ from $t=3295$. Finally, following to the detection procedure, the failure is detected at $t=3305$ with $n=10$. Note that, given the actual time interval of the dataset ($2.5$ mins), this detection is about $3.3$ hours ahead of the actual collapse at this region, allowing a timely intervention for this failure.
  
\section{Conclusion}
 In this study, we introduce $st$-LID, a novel method designed to improve the early and accurate detection of failures in landslide-prone areas. Unlike existing approaches that typically focus on either spatial or temporal aspects of displacement data, $st$-LID effectively incorporate both the spatial correlations and temporal dynamics inherent in the displacement data. By extending the traditional LID framework with a short-term kinematic feature, Bayesian spatial fusion, and long-term temporal modeling, our unified framework robustly identifies the complex spatiotemporal outliers across highly diverse environmental and operational contexts without requiring site-specific tuning. Comprehensive experiments on three distinct real-world datasets demonstrate that $st$-LID provides a robust, interpretable, and computationally efficient foundation for early warning systems. These findings highlight the practical value of $st$-LID in real-world landslide early warning systems, paving the way for timely risk management and potentially saving lives.

\section*{Acknowledgment}
 We would like to thank the GroundProbe for their financial support and provision of data and materials for this study. This research was supported by The University of Melbourne’s Research Computing Services and the Petascale Campus Initiative.

\section*{Generative AI Usage Disclosure}
 The LLM is used at the sentence level (e.g., fixing grammar, re-wording sentences).
\bibliographystyle{splncs04}
\bibliography{ref}

\newpage
\appendix

\section{Displacement Time Series Data}
 We provide visualization of the displacement time series here. Since these datasets are multivariate, for a better visualization, we highlight two time series in-/outside the failure region, and use red dash line to indicate the actual failure time. 
 \begin{figure}[h]
     \centering
     \includegraphics[width=0.65\linewidth]{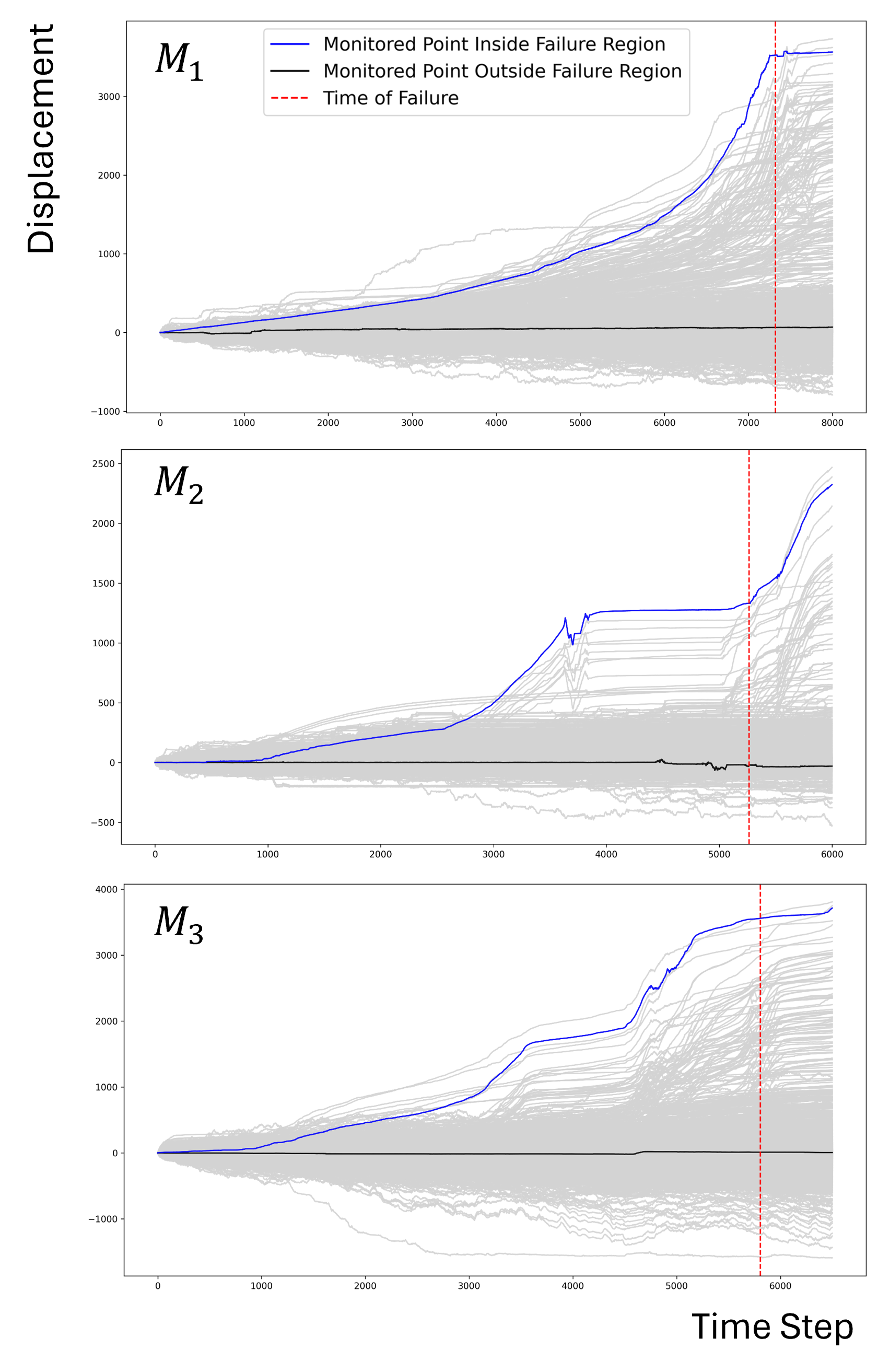}
     \caption{Visualization of the displacement time series.}
     \label{ts_htmp}
 \end{figure}
 Figure \ref{ts_htmp} shows the heatmap of three displacement time series, from which we can see most displacement values stay low all the time with several monitored points experience displacement increasing until their corresponding failure time.


\section{Model Setup}
 We develop the models using a standalone dataset apart from the three datasets used in the main paper. For both the DBSCAN and GCNAE, we utilized a Grid Search strategy. This exhaustive search over a specified subset of the hyper-parameter space allowed us to identify the configurations that maximized model stability and accuracy. 
 \begin{itemize}
     \item DBSCAN: The search focused primarily on the $eps$ (epsilon) and $min\_samples$ parameters. The $eps$ values were tested across a range determined by the k-distance elbow method, while $min\_samples$ was scaled relative to the dimensionality of the input data.

     \item GCNAE: The grid search targeted the architectural depth, the learning rate, and the embedding dimension. We evaluated these based on the minimization of the reconstruction loss and the preservation of topological features in the latent space.
 \end{itemize}
 The final settings in the GCNAE are: A GCN layer that compresses input features into a higher-level spatial representation with hidden dimension of 128. A symmetric GCN structure that reconstructs the original feature dimensions from the hidden state.

 For the intrinsic dimensionality calculations ($s$-LID and $st$-LID), we defined the neighborhood size $k$ as the total population of the dataset (thus, $k=N-1$). This global neighborhood approach ensures that the LID calculation for each node is grounded in the entire manifold's density, providing a stable metric for detecting local deviations (anomalies) against the global distribution.

\section{Visualizations}
\begin{figure}[h]
     \centering
     \includegraphics[width=0.99\linewidth]{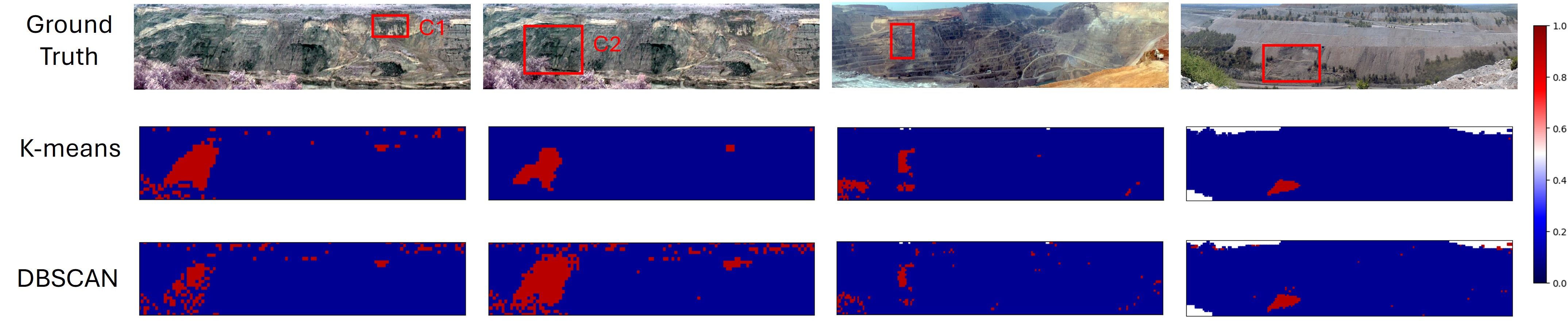}
     \caption{Visualization results of K-means and DBSCAN}
     \label{fig:vis}
 \end{figure}

 Figure \ref{fig:vis} provide visualization results of K-means and DBSCAN.

\end{document}

%% file: math_commands.tex
\usepackage{amsmath,amsfonts,bm}

\def\eqref#1{equation~\ref{#1}}

\def\1{\bm{1}}

\def\vw{{\bm{w}}}
\def\vx{{\bm{x}}}
\def\vy{{\bm{y}}}

\DeclareMathAlphabet{\mathsfit}{\encodingdefault}{\sfdefault}{m}{sl}
\SetMathAlphabet{\mathsfit}{bold}{\encodingdefault}{\sfdefault}{bx}{n}

